\documentclass[10pt,twocolumn,letterpaper]{article}

\usepackage[accsupp]{axessibility}  % Improves PDF readability for those with disabilities.
\usepackage{iccv}
\usepackage{times}
\usepackage{epsfig}
\usepackage{graphicx}
\usepackage{amsmath}
\usepackage{amssymb}
\usepackage{autobreak}
\usepackage{algorithm}
\usepackage{algorithmic}
\usepackage{multirow}
\usepackage{subfigure}
\usepackage{afterpage}
\usepackage{multicol}
\usepackage{caption}
\usepackage[pagebackref=true,breaklinks=true,letterpaper=true,colorlinks,bookmarks=false]{hyperref}

\iccvfinalcopy % *** Uncomment this line for the final submission

\def\iccvPaperID{3582} % *** Enter the ICCV Paper ID here

\ificcvfinal\fi

\begin{document}

%\makeatletter
%\renewcommand\AB@affilsepx{\hfill \protect\Affilfont}
%\makeatother

\makeatletter
\newcommand\email[2][]%
   {\newaffiltrue\let\AB@blk@and\AB@pand
      \if\relax#1\relax\def\AB@note{\AB@thenote}\else\def\AB@note{\relax}%
        \setcounter{Maxaffil}{0}\fi
      \begingroup
        \let\protect\@unexpandable@protect
        \def\thanks{\protect\thanks}\def\footnote{\protect\footnote}%
        \@temptokena=\expandafter{\AB@authors}%
        {\def\\{\protect\\\protect\Affilfont}\xdef\AB@temp{#2}}%
         \xdef\AB@authors{\the\@temptokena\AB@las\AB@au@str
         \protect\\[\affilsep]\protect\Affilfont\AB@temp}%
         \gdef\AB@las{}\gdef\AB@au@str{}%
        {\def\\{, \ignorespaces}\xdef\AB@temp{#2}}%
        \@temptokena=\expandafter{\AB@affillist}%
        \xdef\AB@affillist{\the\@temptokena \AB@affilsep
          \AB@affilnote{}\protect\Affilfont\AB@temp}%
      \endgroup
       \let\AB@affilsep\AB@affilsepx
}
\makeatother

%%%%%%%%% TITLE
\title{Encoder-decoder with Multi-level Attention for 3D Human Shape and Pose Estimation}

%\author{
%Ziniu Wan\thanks{Equal Contribution.}\\
%UCAS\\
%{\tt\small wanziniu19@mails.ucas.ac.cn}
%\and
%Zhengjia Li\footnotemark[1] \\
%Tongji University\\
%{\tt\small zjli1997@tongji.edu.cn}

%\and
%Maoqing Tian\\
%SenseTime Group Limited\\
%{\tt\small tianmaoqing@sensetime.com}

%\and
%Jianbo Liu\\
%Chinese university of Hong Kong\\
%{\tt\small liujianbo@link.cuhk.edu.hk}

%\and
%Shuai Yi\\
%SenseTime Group Limited\\
%{\tt\small yishuai@sensetime.com}

%\and
%Hongsheng Li \\
%Chinese University of Hong Kong\\
%{\tt\small hsli@ee.cuhk.edu.hk}
%}

%\author[1]{Ziniu Wan}
%\author[2]{Zhengjia Li}
%\author[3]{Maoqing Tian}
%\author[4]{Jianbo Liu}
%\author[3]{Shuai Yi}
%\author[4]{Hongsheng Li}
%\affil[1]{University of Chinese Academy of Sciences}
%\email{\url{wanziniu19@mails.ucas.ac.cn}}
%\affil[2]{Tongji University}
%\email{zjli1997@tongji.edu.cn}
%\affil[3]{SenseTime Group Limited}
%\email{\{tianmaoqing,yishuai\}@sensetime.com}
%\affil[4]{Chinese University of Hong Kong}
%\email{liujianbo@link.cuhk.edu.hk, hsli@ee.cuhk.edu.hk}

\author{
Ziniu Wan$^{1}$\thanks{Equal Contribution.}\qquad
Zhengjia Li$^{2}$\footnotemark[1]\qquad
Maoqing Tian$^{3}$\qquad
Jianbo Liu$^{4}$\qquad
Shuai Yi$^{3}$\qquad
Hongsheng Li$^{4}$\\
%$^{1}$ University of Chinese Academy of Sciences\qquad
$^{1}$ Carnegie Mellon University\qquad
$^{2}$ Tongji University\\
$^{3}$ SenseTime Research\qquad
$^{4}$ Chinese University of Hong Kong \\
{\tt\small 
%wanziniu19@mails.ucas.edu.cn\quad
ziniuwan@andrew.cmu.edu\quad
zjli1997@tongji.edu.cn\quad
tianmaoqing@sensetime.com}\\
{\tt\small 
liujianbo@link.cuhk.edu.hk\quad
yishuai@sensetime.com\quad
hsli@ee.cuhk.edu.hk}
}

%\makeatletter
%\let\@oldmaketitle\@maketitle% Store \@maketitle
%\renewcommand{\@maketitle}{\@oldmaketitle% Update \@maketitle to insert...
%  \centering
%  \includegraphics[width=0.95\linewidth]
%    {figs/overview_ste_ktd_v1.pdf}
%  \captionof{figure}{(a) Spatial-temporal attention: In current frame, the color of each pixel represents the spatial attention score, visualizing the importance of the spatial position. The color on the time axis represents the temporal attention score, visualizing the similarity between the corresponding frame and current one. Warmer color indicates higher attention score. (b) Kinematic tree-based hierarchical regression: Our model pays more attention to joints denoted by dots with warmer color.}
%   The direction of the arrows indicates the order of joints generation.}
%  \label{fig:title}
%  \bigskip
% }
%\makeatother

\maketitle
\ificcvfinal\thispagestyle{empty}\fi

\begin{figure*}[t]
\begin{center}
  \includegraphics[width=0.9\linewidth]{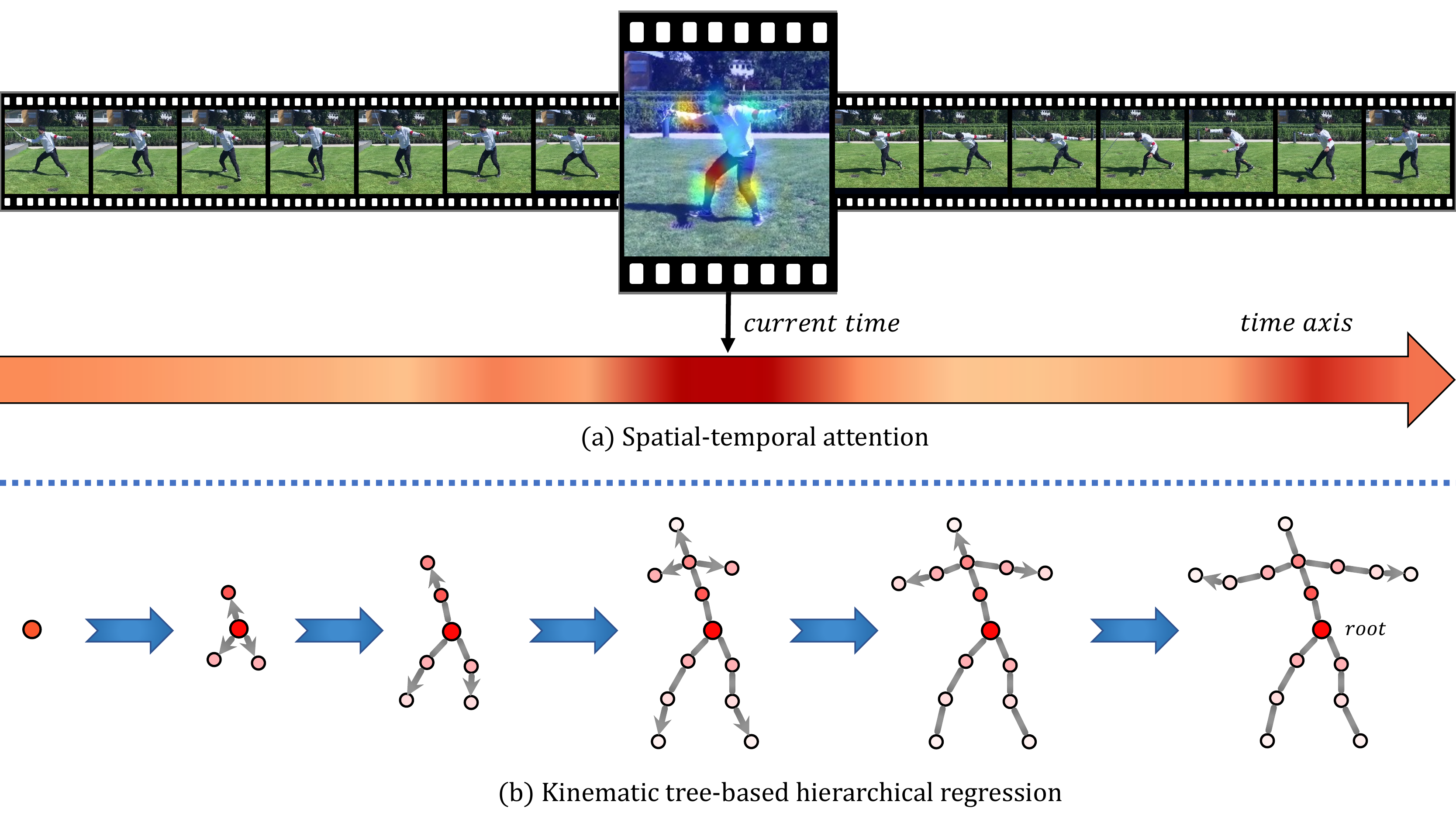}
\end{center}
\caption{(a) Spatial-temporal attention: In current frame, the color of each pixel represents the spatial attention score, visualizing the importance of the spatial position. The color on the time axis represents the temporal attention score, visualizing the similarity between the corresponding frame and current one. Warmer color indicates higher attention score. (b) Kinematic tree-based hierarchical regression: Our model pays more attention to joints denoted by dots with warmer color.}
\vspace{-0.1in}
\label{fig:title}
\end{figure*}

%%%%%%%%% ABSTRACT
\begin{abstract}
3D human shape and pose estimation is the essential task for human motion analysis, which is widely used in many 3D applications. However, existing methods cannot simultaneously capture the relations at multiple levels, including spatial-temporal level and human joint level. Therefore they fail to make accurate predictions in some hard scenarios when there is cluttered background, occlusion, or extreme pose. To this end, we propose Multi-level Attention Encoder-Decoder Network (MAED), including a Spatial-Temporal Encoder (STE) and a Kinematic Topology Decoder (KTD) to model multi-level attentions in a unified framework. STE consists of a series of cascaded blocks based on Multi-Head Self-Attention, and each block uses two parallel branches to learn spatial and temporal attention respectively. Meanwhile, KTD aims at modeling the joint level attention. It regards pose estimation as a top-down hierarchical process similar to SMPL kinematic tree. With the training set of 3DPW, MAED outperforms previous state-of-the-art methods by 6.2, 7.2, and 2.4 mm of PA-MPJPE on the three widely used benchmarks 3DPW, MPI-INF-3DHP, and Human3.6M respectively. Our code is available at \url{https://github.com/ziniuwan/maed}.
\end{abstract}

%%%%%%%%% BODY TEXT
\section{Introduction}

3D human shape and pose estimation from a single image or video is a fundamental topic in computer vision.
% and computer graphics.
% {\color{red}Tremendous works have been made to reduce the reconstruction error and the dependency on multiple sensors (\eg XXX\cite{}, XXX\cite{} and XXX\cite{}).
It is difficult to directly estimate the 3D human shape and pose from monocular images without any 3D information. 
To tackle this problem, massive 3D labeled data and 3D parametric human body models \cite{smpl,smplx,scape} with prior knowledge are necessary.
Tremendous works \cite{hmr,vibe,spin,meva,GraphCMR,I2LMeshNet} based on Deep Neural Network (DNN) have been made to increase the accuracy and robustness of this task. 
% including relations among monocular inputs and the relation derived from prior knowledge of 3D human body. 
% Here we address this problem with an encoder-decoder network, where the encoder learns the spatial-temporal attention from the monocular inputs and the decoder learns the attention at the human joint level based on the prior knowledge of kinematic topology.
% With the emergency of 3D parametric human body model and the accumulation of the massive labeled data, it is possible to use Deep Neural Network (DNN) to predict 3D human shape and pose from a single image or video \cite{hmr,vibe,spin}.

However, existing DNN-based methods often fail in some challenging scenarios, including cluttered background, occlusion and extreme pose. 
To overcome these challenges, three intrinsic relations should be jointly modeled for the video-based 3D human shape and pose estimation:
a). \textit{Spatial relation}: 
%In general computer vision tasks, the CNN is a powerful structures to capture the spatial relations between the different areas in a image.
For the pose estimation task, the human joints areas and the spatial correlations among body parts are directly related to the pose prediction. It is critical to carefully utilize the spatial relation, especially in the scene of cluttered background.
%from different areas of image.  
%Then the ability to focus on the important spatial position in image, which is necessary for cluttered environments. 
b). \textit{Temporal relation}: Everyone has particular temporal trajectory in a given video. In occlusion cases, this temporal relation should be exploited to infer the pose of current occluded frame from surrounding frames.
% for increasing the temporal robustness due to the pose of the current occluded frame can be inferred from the surrounding frames.
c). \textit{Human joint relation}: In the parametric 3D body model SMPL \cite{smpl}, human joints are organized as a kinematic tree. Once pose changes, the parent joint rotates first, and then rotates the children. When the pose amplitude is large, we argue that the prior of the dependence among joints is especially helpful for accurate pose estimation.
% Therefor the bias of the parent joint's estimation incurs substantial negative impact on the estimation of all its children joints.
% The reasonable use of this relation can effectively make an accurate prediction, which is especially useful in the scenario of extreme human pose.
However, none of the existing methods fully utilizes the above three relations in a unified framework.

Motivated by the above observations, we propose Multi-level Attention Encoder-Decoder Network (MAED) for video-based 3D human shape and pose estimation. 
MAED is the first work to explore the above three relations by exploiting corresponding multi-level attentions in a unified framework.
%the spatial-temporal and human joint attentions in a unified framework. 
% MAED is an end-to-end trainable DNN, which includes
It includes Spatial-Temporal Encoder (STE) for spatial-temporal attention and Kinematic Topology Decoder (KTD) for human joint attention.

%these valuable relations at different levels in a unified framework. 
%Specifically, we propose Spatial-Temporal Encoder (STE) which
% The STE is based on Transformer to extract semantic representations from video clip.
% for the subsequent human pose and shape estimation.
Specifically, the STE consists of several cascaded blocks, and each block uses two parallel branches to learn spatial and temporal attention respectively.
We call the two branches Multi-Head Self-Attention Spatial (MSA-S) and Multi-Head Self-Attention Temporal (MSA-T), which are inspired by Multi-Head Self-Attention (MSA) mechanism in Transformer related works \cite{transformer,bert,vit,ipt,detr}.
% Inspired by Multi-Head Self-Attention (MSA) mechanism in Transformer related works \cite{transformer,bert,vit,ipt,detr}, we construct Multi-Head Spatial Self-Attention (MSA-S) module and Multi-Head Temporal Self-Attention (MSA-T) module to capture spatial and temporal relations respectively.
Derived from MSA, MSA-S and MSA-T have Transformer-like structures, but are different in the order of input features dimensions.
As illustrated in Figure \ref{fig:title}(a), MSA-S focuses on the critical spatial positions in image, highlighting significant features for pose estimation.
% and suppressing noisy information in complex outdoor scenarios. 
Meantime, MSA-T concentrates on improving the prediction of current frame by exploiting frames that are informative to current one according to the calculated temporal attention scores.

On the other hand, existing methods usually use an iterative feedback regressor \cite{hmr,vibe} to regress the SMPL \cite{smpl} parameters, in which the pose parameters of all joints are generated simultaneously.
% and the human joint relation is ignored.
However, they ignore the human joint relation.
% dependence between different joints, \textit{i.e.}, the bias of the parent joint's estimation incurs substantial negative impact on the estimation of all its children joints.
% The human joints is organized as a kinematic tree.
% An iterative feedback regressor followed \cite{hmr} is widely used to directly regress the pose parameters $\vec{\theta} \in \mathbb{R}^{K\times3}$ of SMPL\cite{smpl}, with $K=24$ joints, and each joint has 3 parameters. The pose parameters of each joint are estimated simultaneously.
% However, this method ignore the dependence relation between different joints. 
%The root joint 0 controls the global body orientation. Therefore, 
To exploit the dependence among joints, we further propose KTD to simulate the SMPL kinematic tree for joint level attention modeling.
% In KTD, the shape $\vec{\beta}$ and cam $\vec{\phi}$ parameters are still estimated by linear mappings respectively.
% The pose parameters of different joints are not generated at the same time, but in an iterative process from parent joints to children joints in the kinematic tree. 
% The pose parameters of different joints are generated in an iterative process from parent joints to children joints in the kinematic tree asynchronously.
 %As shown in Figure \ref{fig:title}(b), in our proposed KTD, the SMPL parameters are generated by linear regressors. Particularly, the pose parameters are estimated though a top-down hierarchical regression process. To estimate the pose of a joint, except for the image feature, we also take the predicted pose parameters of its ancestors as input of linear regressor. In this way, the pose estimation of child joint is affected by its ancestors, which encourages the model to pay more attention to parent joints, especially the root. The KTD can capture the inherent relation of joints and effectively reduce the prediction error.
% the model captures the inherent sequential relation of joints, and make a more accurate pose estimation with the kinematic topology level attention.
In KTD, each joint is assigned a unique linear regressor to regress its pose parameters. As shown in Figure \ref{fig:title}(b), these parameters are generated through a top-down hierarchical regression process. To estimate a joint, besides image feature, we also take the predicted pose parameters of its ancestors as the input of linear regressor. 
In this manner,
% the biased prediction of ancestor joints will be propagated to child joints,
the bias of the parent joint's estimation incurs substantial negative impact on the estimation of all its children, which forces the KTD to predict more accurate results for ancestor joints.
% the pose estimation of child joint could be affected by its ancestors. 
% Although KTD does not explicitly assign a greater weight to the loss value of the parent joint in the training phase, 
In other words, although KTD does not explicitly allocate an attention score to each joint, the top-down regression process implicitly encourages the model to pay more attention to the parent joints with more children.
% which encourages the model to pay more attention to these parent joints, especially the root. 
As a result, the proposed KTD captures the inherent relation of joints and effectively reduce the prediction error.

We summarize the contributions of our method below:
\begin{itemize}
\item We propose Multi-level Attention Encoder-Decoder Network (MAED) for video-based 3D human shape and pose estimation. Our proposed MAED contains Spatial-Temporal Encoder (STE) and Kinematic Topology Decoder (KTD). It learns different attentions at spatial level, temporal level and human joint level in a unified framework.

%\item We conduct ablation study to quantitatively compare different spatial-temporal attention structures and different regressor structures, proving the effect of our proposed STE and KTD module.

%\item We achieve state-of-the-art results in video-based 3D human shape and pose estimation on many benchmarks.

\item {Our proposed STE leverages the MSA to construct
% into the task of video-based 3D human shape and pose estimation task. Moreover, we propose two modules
MSA-S and MSA-T to encode the spatial and temporal attention respectively in the given video.}

\item {Our proposed KTD
% is one novelty linear regressor for the task of 3D human shape and pose estimation, which
considers hierarchical dependence among human joints and implicitly captures human joint level attention.}
% and yields more robust and accurate prediction.}

\end{itemize}

%------------------------------------------------------------------------
\section{Related Works}

\begin{figure*}
\begin{center}
  \includegraphics[width=0.95\linewidth]{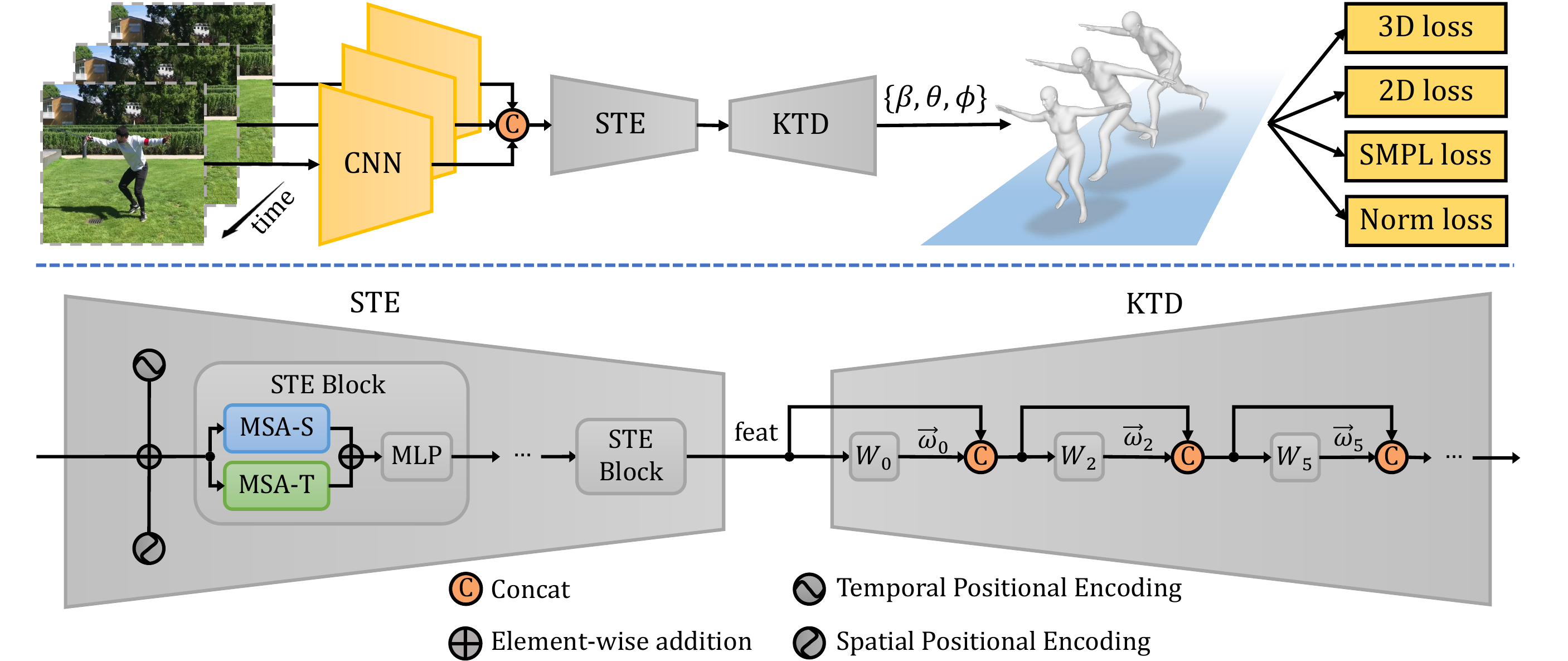}
\end{center}
  \caption{Overview of the proposed method MAED. The upper part shows the pipeline of the model and the lower part presents the structures of our proposed Spatial-Temporal Encoder and Kinematic Topology Decoder.}
\label{fig:net}
\end{figure*}

%-------------------------------------------------------------------------
\subsection{3D Human Shape and Pose Estimation}

% The parametric model
Recent works have made significant advances in 3D human pose and shape estimation due to the parametric 3D human body models, such as SMPL \cite{smpl}, SMPL-X \cite{smplx} and SCAPE \cite{scape}, which
% These models
% parametric 3D human body
utilize the statistics of human body and provide 3D mesh based on few hyper-parameters.
%\textcolor{red}{Based on these parametric 3D human body models, the 3D human shape and pose estimation problem is transformed to regress 3D human models hyper-parameters.}
Later, various studies focus on estimating the hyper-parameters of 3D human model directly from image or video input.
% These approaches predict the hyper-parameters of the specific parametric 3D human body model from image or video inputs. 
%and the 3D human shape and pose results are generated by the parametric 3D human body model. The approaches which directly generate the 3D human shape and pose are not considered in this paper.

% Optimization-based methods and regression-based methods
Previous parametric 3D human body model based methods are split into two categories: optimization-based methods and regression-based methods. The optimization-based methods fit the parametric 3D human body models to pseudo labels, like 2D keypoints, silhouettes and semantic mask. SMPLify \cite{smpl}, one of the first end-to-end optimization-based methods, uses strong statistics priors to guide the optimization supervised by 2D keypoints.
% Besides SMPLify \cite{smpl},
The work \cite{lassner2017unite} utilizes silhouettes along with 2D keypoints to supervise the optimization. On the other hand, regression-based methods train deep neural network to regress the hyper-parameters directly. HMR \cite{hmr} is trained with the supervision of re-projection keypoints loss along with adversarial learning of human shape and pose. SPIN \cite{spin} exploits SMPLify \cite{smpl} in the training loop to provide more supervision. VIBE \cite{vibe} is a video-based method that employs adversarial learning of the motions.

% ----------------------------------------------------------------
\subsection{Transformer in Computer Vision}

Transformer \cite{transformer} is first proposed in NLP field. It is an encoder-decoder model, completely replacing commonly used recurrent neural networks with Multi-Head Self-Attention mechanism, and later achieves great success in various NLP tasks \cite{bert, gpt1, gpt2, gpt3, xlm, bart}.
%that integrates multiple NLP task datasets. 
%This proves that Transformer generalizes well when the amount of training data is sufficient. 
Motivated by the achievements of Transformer in NLP, various works start to apply Transformer to computer vision tasks. Vision Transformer (ViT) \cite{vit} views an image as a 16x16 patch sequence, and trains a
% convolution-free
Transformer for image classification.
% achieving results comparable to the state-of-the-art.
The work \cite{deit} explores distillation to use smaller datasets to get more efficient ViT. Some works \cite{t2tvit,botnet} study various Transformer structures which are more suitable for visual classification tasks. In addition, Transformer has also achieved impressive results in many downstream computer vision tasks, including denoising \cite{ipt}, object detection \cite{detr,ddetr}, video action recognition \cite{vid-trans}, 3D mesh reconstruction \cite{point}, panoptic segmentation \cite{axial}, \textit{etc}.
%It is worth noting that, \cite{metro} uses Transformer to model the relation between image and 3D human mesh vertices for pose estimation and mesh reconstruction.
% It is worth noting that, \cite{metro} uses Transformers for pose estimation and mesh reconstruction.
% It leverages Transformer to learn the relations among 3D human mesh vertices and 3D human joints.
In this paper, we focus on using Transformer to fully exploit the spatial-temporal level attention from video for better human pose and shape estimation.

% \subsection{Human Kinematic Structure for Pose Estimation}
% Learning accurate object pose is inherently difficult because the pose is high dimensional and has many structural constraints. To increase the accuracy and geometric validity.

%However, it leverages only a convolution network to extract spatial feature from image.

%------------------------------------------------------------------------
\section{Methods}

\begin{figure*}[t]
\begin{center}
  \includegraphics[width=\linewidth]{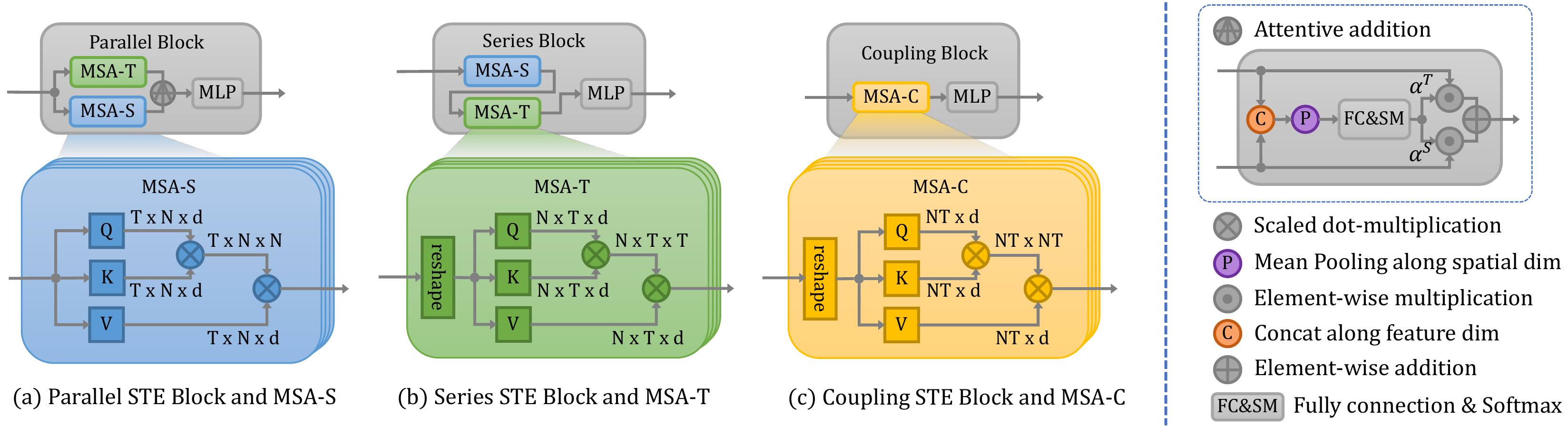}
\end{center}
  \caption{STE block variants and MSA variants.}
\label{fig:ste_blocks}
\end{figure*}

In this section, we first revisit the parametric 3D human body model (SMPL \cite{smpl}). Secondly, we give an overview of our proposed framework.
% which includes one CNN backbone, the Spatial-Temporal Encoder (STE) and the Kinematic Topology Decoder (KTD).
% including network structure and training losses.
Finally, we describe the proposed STE and KTD in detail.
% Finally, we give the necessary training details.

\subsection{SMPL}

% 结合了SPIN和VIBE对于SMPL的描述
SMPL \cite{smpl} is a classical parametric human body model with $N=6890$ vertices and $K=23$ joints. It provides a function $\mathcal{M}(\vec{\beta}, \vec{\theta})$ that takes as input the shape parameters $\vec{\beta} \in \mathbb{R}^{10}$ and the pose parameters $\vec{\theta} \in \mathbb{R}^{72}$, and returns the body mesh $M \in \mathbb{R}^{N\times3}$. $\vec{\beta}$ are the first 10 coefficients of a PCA shape space, controlling the shape of the body (\eg, height, weight, etc). $\vec{\theta}=\left[\vec{\omega}_{0}^{T}, \ldots, \vec{\omega}_{K}^{T}\right]^{T}$ controls the pose of the body, where $\vec{\omega}_{k} \in \mathbb{R}^3$ denotes the axis-angle representation of the relative rotation of joint $k$ with respect to its parent in the kinematic tree. $\vec{\theta}$ is defined by $|\vec{\theta}|=3 \times 23+3=72$ parameters, \textit{i.e.}, 3 for each joint plus 3 for the root orientation. These joints can be calculated by a linear regressor $J_{reg}$, \textit{i.e.}, $J_{3d} \in \mathbb{R}^{K\times3}=J_{reg}M$.

\subsection{Framework Overview}

%Figure \ref{fig:net} shows the structure of our baseline. It takes as input an image, and adopts a ResNet-50\cite{resnet} to extract the basic feature. Following \cite{hmr}, an iterative feedback regressor is used to estimate shape $\vec{\beta}$, pose $\vec{\theta}$ and camera $\vec{\phi}$ parameters. At each iteration, the regressor takes the image feature and current parameter combination $\Theta_t=\{\vec{\beta},\vec{\theta},\vec{\phi}\}$ as input, and outputs the residual $\Delta \Theta_{t}$. The parameter combination is updated by $\Theta_{t+1}=\Theta_{t}+\Delta \Theta_{t}$. The initial $\Theta$ is set to statistical mean $\bar{\Theta}$. These parameters allow us to calculate 3D joints using SMPL model and estimate the 2D projection of these joints, $J_{2d}=\Pi_{\vec{\phi}}\left(J_{3d}\right)$, in which $\Pi_{\vec{\phi}}(.)$ is the projection function.

Figure \ref{fig:net} shows the architecture of our proposed network. It takes a video clip of length $T$ as input, and adopts a CNN backbone to extract the basic feature for each frame. The global pooling layer at the end of the CNN is omitted, resulting in $T$ feature maps of size $(h \times w \times d)$, where $h$/$w$/$d$ denotes the height/width/channel size of feature map. We reshape each feature map into 1D sequence of size $(hw\times d)$, and prepend a trainable embedding to each sequence (Following \cite{vit}, we denote a token in the sequence as a patch). Thus, the CNN outputs a matrix of size $(T\times N \times d)$, where $N=hw+1$. Then our proposed Spatial-Temporal Encoder (STE) is used to perform spatial-temporal modeling on these basic features. The encoded vector corresponding to the prepended embedding serves as the output of STE. Finally, our proposed Kinematic Topology Decoder (KTD) is employed to estimate shape $\vec{\beta}$, pose $\vec{\theta}$ and camera $\vec{\phi}$ parameters from the output of STE.
% maybe briefly explaining KTD here.
These predicted parameters allow us to utilize SMPL to calculate 3D joints and their 2D projection, $J_{2d}=\Pi_{\vec{\phi}}\left(J_{3d}\right)$, where $\Pi_{\vec{\phi}}(.)$ is the projection function.

After getting $\{\vec{\beta},\vec{\theta},J_{3d},J_{2d}\}$, the model is supervised by the following 4 losses:
\begin{equation}
    L=L_{3D}+L_{2D}+L_{SMPL}+L_{NORM}
\end{equation}

%where $\lambda_1=600$, $\lambda_2=300$, $\lambda_3=60$, $\lambda_4=0.06$, $\lambda_5=1$, and each term is calculated as:
where $L_{2D}$/$L_{3D}$ denotes the 2D/3D keypoint loss, $L_{SMPL}$ denotes the SMPL parameters loss, and $L_{NORM}$ denotes the L$_2$-Normalization loss. $J_{3dgt},J_{2dgt},\vec{\theta}_{gt},\vec{\beta}_{gt}$ represent the ground truth of $J_{3d},J_{2d},\vec{\theta},\vec{\beta}$ respectively.
% {\color{red}where $L_{3D}$,$L_{2D}$,$L_{SMPL}$ and $L_{NORM}$ denotes the 3D loss, 2D loss, SMPL loss and norm loss respectively. And these terms are calculated as:}

\begin{equation}
\begin{gathered}
L_{3D} =\sum_{k=1}^{K}\left\|J_{3d}^{k}-J_{3dgt}^{k}\right\|_{2}, \\ 
L_{2D} =\sum_{k=1}^{K}\left\|J_{2d}^{k}-J_{2dgt}^{k}\right\|_{2} \\
L_{SMPL} = \|\vec{\theta}-\vec{\theta}_{gt}\|_{2} + \|\vec{\beta}-\vec{\beta}_{gt}\|_{2} \\
L_{NORM} = \|\vec{\beta}\|_{2}+\|\vec{\theta}\|_{2}
\end{gathered}
\end{equation}

\subsection{Spatial-Temporal Encoder}

Transformer \cite{transformer} is able to effectively model the interaction of tokens in a sequence. %Since a video clip is a sequence of frames, an intuitive way to apply the Transformer to a video is to pool the feature of each frame into a 1-D vector, and take the 1-D vector as a token.
Recently, applying Transformer to model the temporal attention on global pooling feature of each frame is widely used in many video-based computer vision tasks.
% Although this method captures temporal information into the extracted features, 
However, the global pooling operation will inevitably lose the spatial information in a frame, which makes it difficult to estimate detailed human pose.
In our method, to perform spatial and temporal modeling simultaneously, we serialize the input video clip in multiple ways, and design three variants based on Multi-Head Self-Attention (MSA) \cite{transformer}: Multi-Head Spatial Self-Attention (MSA-S), Multi-Head Temporal Self-Attention (MSA-T) and Multi-Head Self-Attention Coupling (MSA-C).
% To further explore the potential of different MSA variants,
Then we further design three forms of Spatial-Temporal Encoder (STE) Block as shown in Figure \ref{fig:ste_blocks}, which endows the encoder with both global spatial perception and temporal reasoning capability. Finally, we stack multiple STE Blocks to construct the STE.
% In this section, we mainly discuss the structure design of the different MSA and STE Block variants, and in Section4, we will show the quantitative results.

\textbf{MSA Variants}. The standard MSA can only learn attention of one dimension, so the different order of input dimensions affects the meaning of learned attention. Our proposed three variants have similar model structure, but are different on the order of the input dimensions.
% The MSA variants takes three dimension features maps as input, and the output feature maps have the same shape as input. 
% The input feature maps is a matrix of size $(T \times (N + 1) \times d)$ corresponding to one video clip,  where $T$ is the length of video clip, $N=H'\times W'$, and $d$ represents feature size. 
% Inspired by \cite{vit}, we embed RGB frames into basic vision representations using a modified ResNet-50\cite{resnet} with the last global pooling layer omitted, and reshape the output 2D feature maps of size $(H',W')$ into 1D sequences of length $N$. 

% Inspired by \cite{vit}, we first embed RGB frames into basic vision representation using a modified ResNet-50\cite{resnet} with the last global pooling layer omitted, and reshape the output 2D feature maps of size $(H',W')$ into 1D sequences. Following \cite{vit}, we denote a token in the sequence as a patch. Each patch represents the feature of a region in a frame. Thus, a input video clip can be represented by a matrix of size $(T \times N \times d)$, where $T$ is the length of video clip, $N=H'\times W'$, and $d$ represents feature size. 
% By this way, the STE Block combined with multiple different MSA variants have ability to capture the spatial-temporal attention simultaneously.

MSA-S aims at finding the key spatial information in a frame, such as joints and limbs of human body.
% and extract representation with rich information. 
%Then, the input video clip represented by a matrix $\in \mathbb{R}^{T \times N \times d}$ will be fed to 
It is shown in the blue box in Figure \ref{fig:ste_blocks}(a), where each self-attention head outputs a heatmap of size $(T \times N \times N)$
%$\mathbf{H}^S \in \mathbb{R}^{T \times N \times N}$ 
computed by scaled dot-multiplication. However, in this setting, temporal relations among frames are not captured, as a patch in one frame does not interact with any patch in other frames.

MSA-T is pretty similar to MSA-S, except that it first reshapes the input matrix from size $(T \times N \times d)$ to $(N \times T \times d)$
% and then use it to compute scaled dot-product attention, 
as shown in the green box in Figure \ref{fig:ste_blocks}(b).
% Different from MSA-S,
Each head of MSA-T outputs the heatmap of size $(N \times T \times T)$
%$\mathbf{H}^T \in \mathbb{R}^{N \times T \times T}$
, where each score reflects the attention of a patch to the patch in the same position in other frames. Although temporal semantics is modeled explicitly, MSA-T ignores spatial relation of patches in the same frame.

% Multi-Head Coupling Self-Attention (MSA-C). 
% MSA-S and MSA-T capture spatial-temporal information in a separated fashion, while
MSA-C flattens patch sequence and frame sequence together, \textit{i.e.}, reshape the input matrix from size $(T \times N \times d)$ to $(TN \times d)$, as shown in the yellow box in Figure \ref{fig:ste_blocks}(c). In this way, the heatmap of size $(TN \times TN)$
%$\mathbf{H}^C \in \mathbb{R}^{TN \times TN}$ 
% computed by scaled dot-multiplication
enables each patch interacts with any other patches in the video clip. 

\textbf{STE Blocks.} As depicted in Figure \ref{fig:ste_blocks}, we design three kinds of STE Blocks based on these MSA variants.
% denoted as Parallel Block, Series Block and Coupling Block.
%Parallel Block and Series Block connect MSA-S and MSA-T in parallel and in series respectively, while Coupling Block only use MSA-C. MSA-S and MSA-T in the same block share parameters to reduce model size.
Coupling Block consists of a MSA-C followed by a Multi-Layer Perception (MLP) layer, modeling spatial-temporal information in a coupling fashion. However, it greatly increases the complexity since the complexity of dot-multiplication is quadratic to sequence length. 

Parallel Block and Series Block connect MSA-S and MSA-T in parallel and in series respectively. For Parallel Block, a naive way of integrating two branches is to simply compute the element-wise mean of the outputs of MSA-S and MSA-T. In order to dynamically balance the temporal and spatial information, we compute attentive weights $\alpha^S,\alpha^T \in \mathbb{R}^{T\times 1 \times d}$ for the two branches. They represent attention scores for the temporal and spatial component along the feature channels of each frame respectively.

Connection of MSA-T and MSA-S makes it possible to combine image and video datasets to train more robust models. When it comes to image input, we simply bypass or disconnect the MSA-T in the blocks to ignore the non-existent temporal information.
% Self-Attention accepts variable length sequences as input, making it possible for us to combine video datasets with image datasets to train more robust models.

Considering the trade-off between accuracy and speed, we empirically choose Parallel Block in our STE, as the Parallel Block is able to dynamically adjust the attentive weights between spatial and temporal attention and yields the best results compared with other variants. The quantitative comparison is discussed in Section \ref{ablation_ste} in detail.

%\textbf{Temporal Positional Encoding.} Since all frames use shared Positional Encoding in Eq\ref{eq:embedding}, we additionally add Temporal Positional Encoding to the input to inject information about the temporal position of frames in the video clip. Thus, Eq\ref{eq:embedding} becomes:
%\begin{align}
%\mathbf{z}_t^0 &= E([\mathbf{x}_{cls};\mathbf{P}_t])+\mathbf{E}^{temp}_t
%\end{align}
%Here, $\mathbf{E}^{temp}_t \in \mathbb{R}^{1 \times D}$ is the $t$th row of learnable Temporal Encoding matrix $\mathbf{E}^{temp} \in \mathbb{R}^{N \times D}$. $\mathbf{E}^{temp}_t \in \mathbb{R}^{1 \times D}$ will be broadcast to all the patches of $t$th frame. 

\textbf{Spatial-Temporal Positional Encoding.} In order to locate the spatial and temporal position of a patch, we add two separated positional encodings to inject sequence information into the input, namely spatial positional encoding $\mathbf{E}^{S}_{pos} \in \mathbb{R}^{1 \times N \times d}$ and temporal positional encoding $\mathbf{E}^{T}_{pos} \in \mathbb{R}^{T \times 1 \times d}$. They are both trainable and added to the input sequence matrix. 
% Data on the skeleton dimension will be broadcast.

\subsection{Kinematic Topology Decoder}

\begin{table*}[]
\begin{center}
\resizebox{\textwidth}{!}{
\begin{tabular}{|l|l|cccc|cc|cc|}
\hline
\multirow{2}{*}{Models}                         & \multirow{2}{*}{Input}  & \multicolumn{4}{c|}{3DPW}                                                                                       & \multicolumn{2}{c|}{MPI-INF-3DHP}                         & \multicolumn{2}{c|}{Human3.6M}                             \\ \cline{3-10} 
                                                &                         & \multicolumn{1}{l}{PA-MPJPE} & \multicolumn{1}{l}{MPJPE} & \multicolumn{1}{l}{PVE} & \multicolumn{1}{l|}{ACCEL} & \multicolumn{1}{l}{PA-MPJPE} & \multicolumn{1}{l|}{MPJPE} & \multicolumn{1}{l}{PA-MPJPE} & \multicolumn{1}{l|}{MPJPE}  \\ \hline \hline
HMR\cite{hmr} w/o 3DPW                          & image                   & 81.3                         & 130.0                     & -                       & 37.4                       & 89.8                         & 124.2                      & 56.8                          & 88.0                       \\
GraphCMR\cite{GraphCMR} w/o 3DPW                & image                   & 70.2                         & -                         & -                       & -                          & -                            & -                          & 50.1                          & -                          \\
STRAPS\cite{STRAPS} w/ 3DPW                     & image                   & 66.8                         & -                         & -                       & -                          & -                            & -                          & 55.4                          & -                          \\
ExPose\cite{ExPose} w/o 3DPW                    & image                   & 60.7                         & 93.4                      & -                       & -                          & -                            & -                          & -                             & -                          \\
SPIN\cite{spin} w/o 3DPW                        & image                   & 59.2                         & 96.9                      & 116.4                   & 29.8                       & 67.5                         & 105.2                      & 41.1                          & -                          \\
I2LMeshNet\cite{I2LMeshNet} w/o 3DPW            & image                   & 57.7                         & 93.2                      & -                       & -                          & -                            & -                          & 41.1                          & \textbf{55.7}              \\
Pose2Mesh\cite{Pose2Mesh} w/o 3DPW              & 2D Pose                 & 58.3                         & 88.9                      & -                       & -                          & -                            & -                          & 46.3                          & 64.9                       \\
TemporalContext\cite{temporal_context} w/o 3DPW & video                   & 72.2                         & -                         & -                       & -                          & -                            & -                          & 54.3                          & 77.8                       \\
DSD-SATN\cite{dsd_satn} w/o 3DPW                & video                   & 69.5                         & -                         & -                       & -                          & -                            & -                          & 42.4                          & 59.1                       \\
MEVA\cite{meva} w/ 3DPW                         & video                   & 54.7                         & 86.9                      & -                       & \textbf{11.6}              & 65.4                         & 96.4                       & 53.2                          & 76.0                       \\
VIBE\cite{vibe} w/o 3DPW                        & video                   & 56.5                         & 93.5                      & 113.4                   & 27.1                       & 63.4                         & 97.7                       & 41.5                          & 65.9                       \\
VIBE\cite{vibe} w/ 3DPW                         & video                   & 51.9                         & 82.9                      & 99.1                    & 23.4                       & 64.6                         & 96.6                       & 41.4                          & 65.6                       \\ \hline
Ours w/o 3DPW                                   & video                   & \textbf{50.7}                & \textbf{88.8}             & \textbf{104.5}          & 18.0                       & \textbf{56.5}                & \textbf{85.1}              & \textbf{38.7}                 & 56.3                       \\ 
Ours w/ 3DPW                                    & video                   & \textbf{45.7}                & \textbf{79.1}             & \textbf{92.6}           & 17.6                       & \textbf{56.2}                & \textbf{83.6}              & \textbf{38.7}                 & 56.4                       \\ \hline
\end{tabular}
}
\end{center}
\caption{Performance comparison with the state-of-the-art methods on 3DPW, MPI-INF-3DHP and Human3.6M datasets. The bold font represents the best result.}
\label{fig:compared_with_sota}
\end{table*}

% Previous methods usually adopt an iterative feedback regressor \cite{vibe,hmr} to regress the pose parameter of all joints simultaneously, which
As aforementioned, previous works ignore the inherent dependence among joints and regard them as equally important. Therefore, we design Kinematic Topology Decoder (KTD) to implicitly model the attention at the joint level.
% to focus more on the optimization of parent joints.
 
\begin{figure}[t]
\begin{center}
  \includegraphics[width=0.7\linewidth]{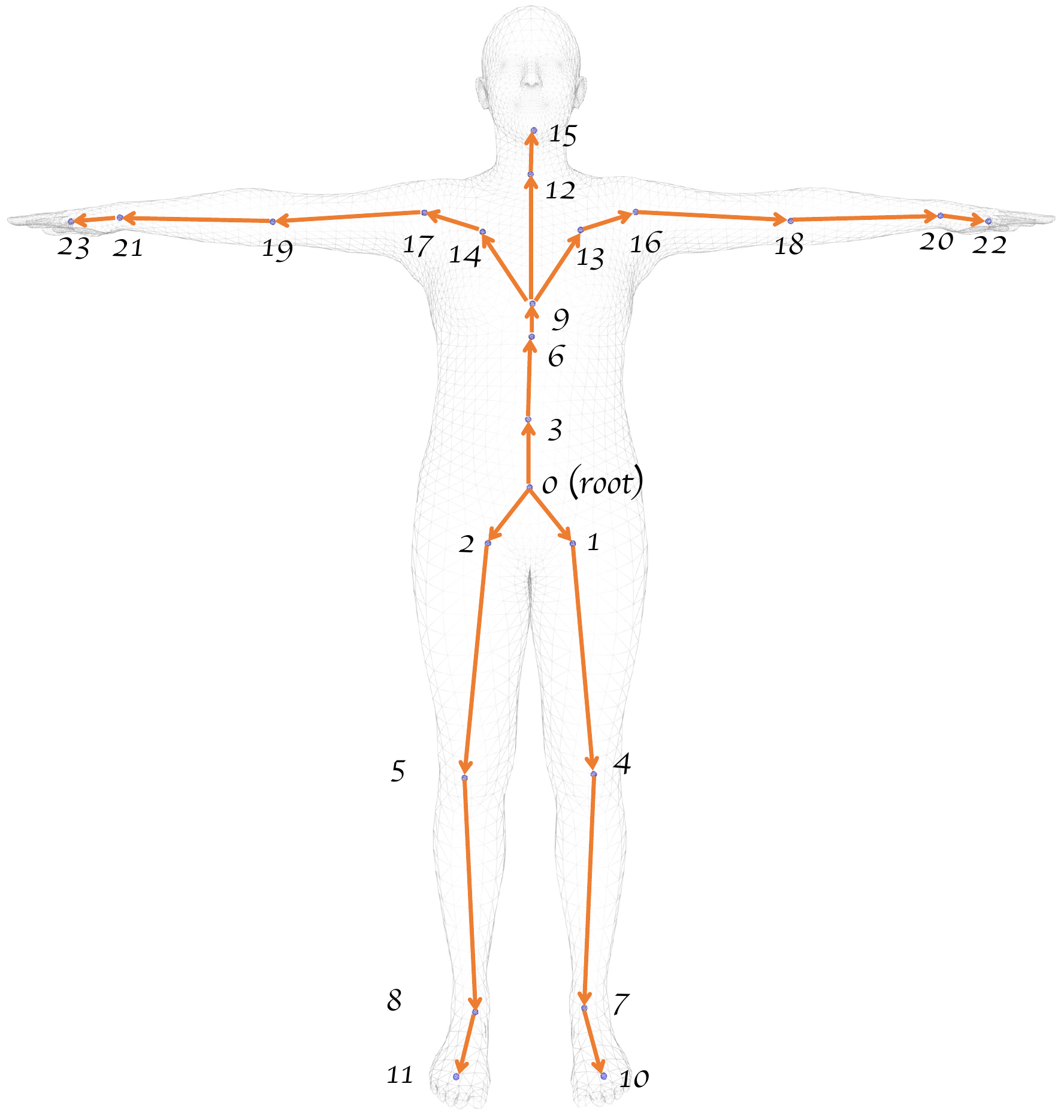}
\end{center}
  \vspace{-0.1in}
  \caption{A demonstration of kinematic tree with 23 joints and a root. The arrow points from the parent to its child.}
\label{fig:body_with_joints}
\end{figure}

As demonstrated in Figure \ref{fig:body_with_joints}, the pose of human body is controlled by 23 joints which are organized as a kinematic tree.
% The arrow points from the parent joint to its child, and we can always reach to any joint from the root joint following a path in the tree.
We first revisit how the pose parameters rotate the joints in SMPL \cite{smpl}. As shown in Eq (\ref{equ:transformation}), the world transformation of joint $k$ denoted by $G_{k}(\mathcal{R}, \mathcal{T}) \in \mathbb{R}^{4\times4}$ equals the cumulative product of the transformation matrices of its ancestors in the kinematic tree.

\begin{equation}
    G_{k}(\mathcal{R}, \mathcal{T})=\prod_{i \in A(k)}\left(\begin{array}{cc}
\mathbf{R}_i & \mathbf{t}_{i} \\
\mathbf{0} & \mathbf{1}
\end{array}\right)
\label{equ:transformation}
\end{equation}

where $\mathcal{R}=[\mathbf{R}_0,...,\mathbf{R}_K],\mathcal{T}=[\mathbf{t}_0,...,\mathbf{t}_K]$. Following SMPL, $\mathbf{R}_k \in \mathbb{R}^{3\times3}$ and $\mathbf{t}_k \in \mathbb{R}^{3\times1}$ denote the rotation matrix and translation vector of joint $k$ respectively.
% $\mathbf{R}_k$ is transformed from the axis-angle $\vec{\omega}_{k}$ using \textit{Rodrigues formula}.
$A(k)$ is the ordered set of joint $k$'s ancestors, \eg, $A(5)=\{0,2\}$.
% $\mathbf{R}_k$ is transformed from the axis-angle $\vec{\omega}_{k}$ using the \textit{Rodrigues formula} described in Equ (\ref{equ:rod_form}). where $\widehat{\bar{\omega}}_{k}$ is the skew symmetric matrix of $\vec{\omega}_{k}$ and $\mathcal{I}$ is the $3\times3$ identity matrix. $A(k)$ is the ordered set of joint $k$'s ancestors. Take Figure \ref{fig:body_with_joints} as an example, $A(5)=\{0,2\}$.
% $\mathbf{J}$ are the joint locations, and each 3-element vector in $\mathbf{J}$ corresponding to a single joint center, $j$, is denoted $\mathbf{j}_j$.
% \begin{equation}
%     \mathbf{R}_k=\text{rod}(\vec{\omega}_{k})=\mathcal{I}+\widehat{\bar{\omega}}_{k} \sin \left(\left\|\vec{\omega}_{k}\right\|\right)+\widehat{\bar{\omega}}_{k}^{2} \cos \left(\left\|\vec{\omega}_{k}\right\|\right)
% \label{equ:rod_form}
% \end{equation}

Therefore, the position of a joint is affected by its own and ancestral pose parameters. The more children a joint has, the greater its impact on the accuracy of the overall joint position estimation. Despite this, currently widely used iterative feedback regressor \cite{hmr,vibe}
% which regress the pose parameters of all joints simultaneously,
does not pay more attention to the parent joints, especially the root of kinematic tree.
As a result, it can only get sub-optimal results.
However, our proposed KTD can avoid the problem. In KTD, we first decode the shape/cam parameters with a matrix $\mathbf{W}_{\text{shape}}$/$\mathbf{W}_{\text{cam}}$ as shown in Eq (\ref{eq:shape_cam}).
\begin{equation}
    \vec{\beta}=\mathbf{W}_{\text{shape}} \cdot \mathbf{x}, \quad \vec{\phi}=\mathbf{W}_{\text{cam}} \cdot \mathbf{x}
\label{eq:shape_cam}
\end{equation}
% Based on above observations, we conclude the following shortcomings of current widely used iterative feedback regressor, which regress the pose parameters of all joints simultaneously.
% \begin{itemize}
%     \item The parent joints (especially the root joint) are not paid enough attention. The error of their pose estimation will bring a great negative impact on children joints.
%     \item There is a dependence among the pose parameters of different joints, \textit{i.e.}, to calculate the transformation of a joint $k$, we need to estimate the pose parameters of all its ancestors. However, current regressor generate the pose parameters for all joints at the same time, and ignore the dependence among them.
% \end{itemize}
% In our baseline, a Multi Layer Perception (MLP) is used to decode the pose parameters $\vec{\theta}$. However, it regards the pose parameters of all joints as independent of each other, and ignores the topology relation between them. What's more, the position of leaf joint sometimes becomes inaccurate because of the error of root joint. As a result, it can only get a sub-optimal estimation.
where $\mathbf{W}_{\text{shape}} \in \mathbb{R}^{10\times d}$, $\mathbf{W}_{\text{cam}} \in \mathbb{R}^{3\times d}$, and $\mathbf{x} \in \mathbb{R}^{d}$ is the image feature extracted by the STE.

Then we iteratively generate the pose parameter for each joint in hierarchical order according to the structure of kinematic tree.
Take joints \{0, 2, 5\} in Figure \ref{fig:body_with_joints} as an example. We first predict the pose parameters of root, namely the global body orientation, using the output feature of STE and a learnable linear regressor $\mathbf{W}_{0} \in \mathbb{R}^{6\times d}$, \textit{i.e.}, $\vec{\omega}_0=\mathbf{W}_{0} \cdot \mathbf{x}$. Here, following \cite{spin}, we use the 6D rotation representation proposed in \cite{6d_rotation} for faster convergence. Then, for its child joint 2, we take the image feature $\mathbf{x}$ and $\vec{\omega}_0$ as the input of another linear regressor $\mathbf{W}_{2} \in \mathbb{R}^{6\times (d+6)}$ which outputs the pose parameters $\vec{\omega}_2$, \textit{i.e.}, $\vec{\omega}_2=\mathbf{W}_{2} \cdot Concat(\mathbf{x},\vec{\omega}_0)$, where $Concat(\cdot)$ is the concatenate operation. Similarly for the grandson joint 5, $\vec{\omega}_5=\mathbf{W}_{5} \cdot Concat(\mathbf{x},\vec{\omega}_0,\vec{\omega}_2),\mathbf{W}_{5} \in \mathbb{R}^{6\times (d+12)}$. This regression process is shown in Figure \ref{fig:net}.

By KTD, we establish the dependence between the parent joint and its children, which is consistent with kinematic tree structure. In traditional regressor, the error of the parent joint's pose estimation only affects itself. While in KTD, the error will be propagated to its children as well. This encourages the model to learn an attention at the joint level and pay more attention to parent joints, so as to achieve more accurate estimation results.

% \begin{algorithm}[t]
%     \renewcommand{\algorithmicrequire}{\textbf{Input:}}
%     \renewcommand{\algorithmicensure}{\textbf{Output:}}
%     \begin{algorithmic}[1]
%         \REQUIRE ~~\\
%         Image feature, $x$\\
%         Number of joints, $K$\\
%         \ENSURE ~~\\
%         Pose parameter for each joint
%         \STATE $\vec{\theta}=[]$
%         \FOR{each $k$ in $[0,1,,...,K-1]$}
%         \STATE $\text{inp}=x$
%         \FOR{each $parent$ in $A(k)$}
%         \STATE $\text{inp}=Concat(\text{inp},\vec{\omega}_{parent})$
%         \ENDFOR
%         \STATE $\vec{\omega}_k=\mathbf{W}_k.\text{inp}$
%         \STATE Insert $\vec{\omega}_k$ into $\vec{\theta}$
%         \ENDFOR
%         \RETURN $\vec{\theta}$
%     \end{algorithmic}
%     \caption{Skeleton Topology Decoder}
%     \label{algo:Skeleton Topology Decoder}
% \end{algorithm}

%------------------------------------------------------------------------
\section{Experiments}

\subsection{Datasets}

% In this subsection, we first describe the datasets used for training. Next we give a quick description about the evaluation datasets and metrics used in our experiments. 

\textbf{Training.} Following previous works \cite{hmr}\cite{spin}\cite{vibe}, we use mixed datasets for training, including 3D video datasets, 2D video datasets and 2D image datasets. For 3D video datasets, Human3.6M \cite{human3.6m} and MPI-INF-3DHP \cite{mpii3d} provide 3D keypoints and SMPL parameters in indoor scene. For 2D video datasets, PennAction \cite{pennaction} and PoseTrack \cite{posetrack} provide ground-truth 2D keypoints annotation, while InstaVariaty \cite{insta} provides pseudo 2D keypoints annotation using a keypoint detector \cite{detector1,multiposenet}. For image-based datasets, COCO \cite{coco}, MPII \cite{mpii} and LSP-Extended \cite{lspet} are adopted, providing in-the-wild 2D keypoints annotation. Meanwhile, we conduct ablation study on the 3DPW \cite{3dpw} dataset.

\textbf{Evaluation.} We report the experiments results on Human3.6M \cite{human3.6m}, MPI-INF-3DHP \cite{mpii3d} and 3DPW \cite{3dpw} evaluation set. We adopt the widely used evaluation metrics following previous works \cite{hmr}\cite{spin}\cite{vibe}, including Procrustes-Aligned Mean Per Joint Position Error (PA-MPJPE), Mean Per Joint Position Error (MPJPE), Per
Vertex Error (PVE) and ACCELeration error (ACCEL). We report the results with and without 3DPW \cite{3dpw} training set for fair comparison with previous methods.

\subsection{Training Details}

\textbf{Data Augmentation.} Horizontal flipping, random cropping, random erasing \cite{randomerase} and color jittering %with a probability of 0.5, 0.2 and 0.3 respectively 
are employed to augment the training samples. 
%The ratio of the side length of the cropped image to that of the original image is a random number between 0.6 and 1. 
Different frames of the same video input share consistent augmentation parameters. 
% These data augmentations will perform identical operation on each frame of an input video clip to maintain continuity.

% \textbf{Data Augmentation.} Random crop and random horizontal flip are employed to augment training samples. In the random crop technique, we randomly crop a sub-image from the original image with a probability of 0.2. The ratio of the side length of the cropped image to the side length of the original image is a random number between 0.6 and 1. In the random horizontal flip, the input image and its annotation will be randomly flipped horizontally with a probability of 0.5. Different frames of the same video share the data augmentations for continuity. % These data augmentations will perform identical operation on each frame of an input video clip to maintain continuity.

\textbf{Model Details.} Following \cite{vit}, we use a modified ResNet-50 \cite{resnet} as the CNN backbone to extract the basic feature of an input image.
% Specifically, following \cite{vit}, we make three modifications: 1. Replace Batch Normalization with Group Batch Normalization \cite{group_bn}. 2. Remove the fourth stage and increase the number of blocks in the third stage to 9. As a result, the number of blocks per stage changes from [3, 4, 6, 3] to [3, 4, 9]. 3. Remove the global pooling layer. 
%For the STE, a 6-layer Transformer equipped with Spatial-Temporal Self-Attention is concatenated after the ResNet-50.
For STE, 6 STE Parallel Blocks are stacked, and each block has 12 heads. We adopt the weights from \cite{vit} to initialize the ResNet-50 and STE.

The whole training process is divided into two stages. In the first stage, the model aims at accumulating sufficient spatial prior knowledge, and thus is trained with all image-based datasets and frames from Human3.6M and MPI-INF-3DHP.
% and thus is trained with only static-image format using the image-based datasets, the Human3.6M dataset and the MPI-INF-3DHP dataset.
% only static-image format using the image-based datasets, the Human3.6M dataset and the MPI-INF-3DHP dataset. 
We fix the number of epochs as 100 and the mini-batch size as 512 for this stage. In the second stage, we use both video and image datasets for temporal modeling. For video datasets, 
%we first sample 128 consecutive frames for each video clip. From these 128 frames, we start from a random position and sample a 16-frame clip at a equal interval of 8 frames, that we take as a video training instance. 
we sample 16-frame clips at a interval of 8 as training instances.
We train another 100 epochs with a mini-batch size of 32 for this stage. %For both stages, the MSA-T module in our architecture will be temporarily disconnected when the training samples come from the image datasets. 
The model is optimized by Adam optimizer with an initial learning rate of $10^{-4}$ which is decreased by 10 at the 60-th and 90-th epochs. Finally, each term in the loss function has different weighting coefficients. Refer to Sup. Mat. for further details. All experiments are conducted on 16 Nvidia GTX1080ti GPUs.

% A whole training process is divided into two stages. In the first stage, we aim to make the model accumulate sufficient spatial prior knowledge. To implement this, we disconnect all the MSA-T blocks, and use only image-based datasets along with Human3.6M and MPI-INF-3DHP. We train 100 epochs with a mini-batch of size 512 for this stage. In the second stage, we connect MSA-T to the network and use mixed datasets. For all the video datasets, we first sample 128 consecutive frames for each video clip. From these 128 frames, we start from a random position and sample a 16-frame clip at a equal interval of 8 frames, that we take as a video training instance. In order to make full use of the image datasets, we will temporarily disconnect the MSA-T in the model after every 32 iterations of training using the video datasets, and use the image datasets to train for 1 iteration, and then reconnect MSA-T. We choose to use a batch size of 32 and train for another 100 epochs for this stage. For both stages, we use 16 Nvidia GTX1080ti GPUs with a mini-batch evenly distributed on them. Adam optimizer with an initial learning rate of $10^{-4}$ is adopted, and we reduce the learning rate to 1/10 of the previous at 60 and 90 epochs respectively. Following \cite{vibe}, we use different weighting coefficients for different term in the loss function.

\subsection{Comparison to state-of-the-art results}

In this section, we compare our method with the state-of-the-art models on 3DPW, MPI-INF-3DHP and Human3.6M, and the results are summarized in Table \ref{fig:compared_with_sota}. On the 3DPW and MPI-INF-3DHP datasets, our method outperforms other competitors including image- and video-based methods by a large margin, whether or not using 3DPW training set. On Human3.6M, our method achieves results on-par with I2LMeshNet \cite{I2LMeshNet}. We also observe MEVA \cite{meva}, an two-stage method that aims at producing both smooth and accurate results, ranks best in ACCEL metric on 3DPW. However, considering all indicators, our method achieves better performance overall.
% significant improvement on ACCEL metric due to the temporal modeling capability of Multi-Head Temporal Self-Attention.
% What's more, these video-based methods are better than image-based methods, which is reasonable due to the additional temporal information in video.

These results validate our hypothesis that the exploitation of the attentions at spatial-temporal level and human joint level greatly helps to achieve more accurate estimation. The leading performance on these three datasets (especially the in-the-wild dataset 3DPW) demonstrates the robustness and the potential to real-world applications of our method.

% \textbf{3DPW.} From Table \ref{fig:compared_with_sota}, we can see that our method outperforms other competitors by a large margin. Compared with the state-of-the-art video-based model VIBE, we achieve 6 mm, 3.8 mm, 6.5 mm, 5.8 mm/s$^2$ improvements in PA-MPJPE, MPJPE, PVE, ACCEL respectively.
% This demonstrates the effectiveness of modeling spatial-temporal sequential relation and utilizing skeleton topology structure. We also observe that the performance of video-based methods are better than frame-based methods. This is reasonable because video data contains more temporal information.

% \textbf{MPI-INF-3DHP.} Our model still ranks best among these methods. We observe significant 11.8 mm improvements in the MPJPE metric. It is because our Skeleton Topology Decoder tends to pay more attention to parent joints, which lead to more accurate joint estimation.

% \textbf{Human3.6M.} Although our method is 1.4 mm lower than I2LMeshNet in MPJPE metric, we perform better in PA-MPJPE metric (about 2.1 mm improvements). Therefore our method achieve competitive performance as well.

% The excellent performance on these three datasets (especially the in-the-wild dataset 3DPW) demonstrates the robustness of our method, and the great potential to real-world applications.

\begin{figure*}
\centering     %%% not \center
\subfigure
% [Extreme pose]
{\includegraphics[width=0.95\linewidth]{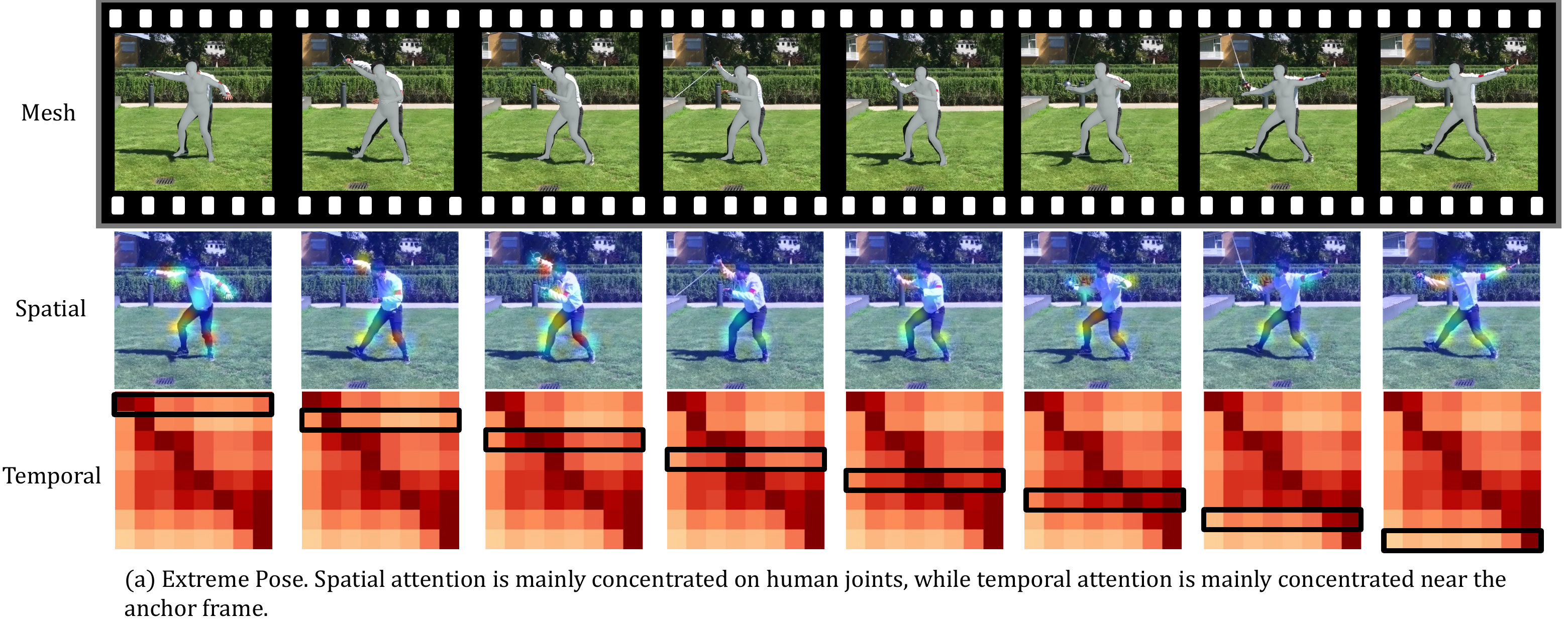}}
\subfigure
% [Back view under cluttered background and occlusion]
{\includegraphics[width=0.95\linewidth]{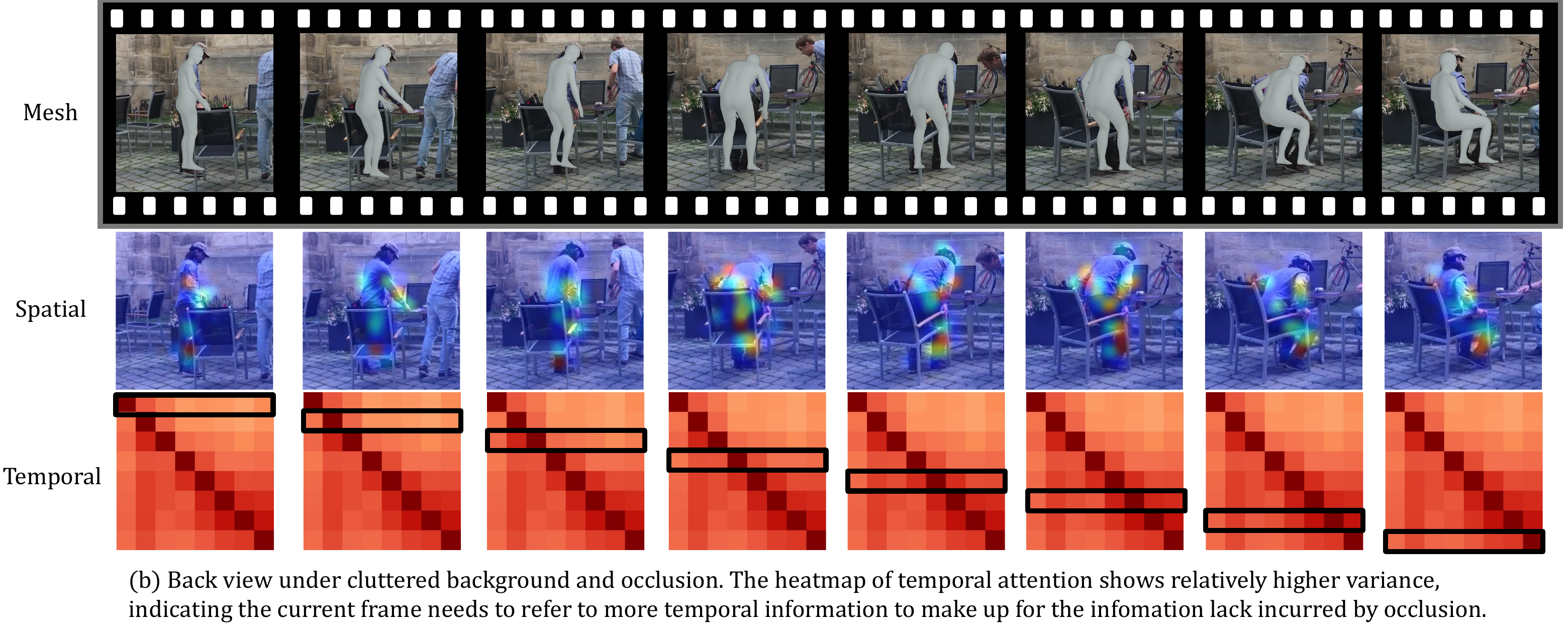}}
\vspace{-0.1in}
\caption{Qualitative visualization of MAED. More visualization results will be shown in Sup.Mat. }
% {First line: Raw image and reconstructed mesh. Second line: Spatial attention map. Third line: Temporal attention map. Each row corresponds to a different frame.}
\label{fig:visual}
\end{figure*}

\subsection{Ablation Study} \label{ablation}

\begin{table}[]
\begin{center}
\resizebox{0.9\columnwidth}{!}{
\begin{tabular}{|l|l|cc|}
\hline
\multirow{2}{*}{Encoder}        & \multirow{2}{*}{Decoder}       & \multicolumn{2}{c|}{3DPW}                                 \\ \cline{3-4} 
                                &                                & \multicolumn{1}{l}{PA-MPJPE} & \multicolumn{1}{l|}{MPJPE} \\ \hline \hline
CNN                             & Iterative                      & 52.2                         & 87.5                       \\
CNN                             & KTD                            & 50.9                         & 88.0                       \\
CNN+STE                         & Iterative                      & 47.5                         & 80.2                       \\
CNN+STE                         & KTD                            & \textbf{45.7}                & \textbf{79.1}              \\ \hline
CNN                             & \multirow{7}{*}{Iterative}     & 52.2                         & 87.5                       \\
CNN+TE                          &                                & 51.1                         & 84.5                       \\
CNN+SE                          &                                & 49.8                         & 84.5                       \\
CNN+STE$_{\text{series}}$       &                                & 48.5                         & 83.6                       \\ 
CNN+STE$_{\text{parallel}}$v1   &                                & 48.1                         & 81.6                       \\
CNN+STE$_{\text{parallel}}$v2   &                                & \textbf{47.5}                & \textbf{80.2}              \\
CNN+STE$_{\text{coupling}}$     &                                & 49.3                         & 82.6                       \\ \hline
\multirow{5}{*}{CNN+STE}        & Iterative                      & 47.5                         & 80.2                       \\
                                & Decoder$_{\text{vanilla}}$     & 47.7                         & 80.7                       \\
                                & KTD                            & \textbf{45.7}                & \textbf{79.1}              \\
                                & KTD$_{\text{random}}$          & 47.7                         & 82.5                       \\
                                & KTD$_{\text{reverse}}$         & 47.6                         & 79.7                       \\ \hline
\end{tabular}
}
\end{center}
\caption{Analytical experiment results with different encoders and decoders. CNN represents ResNet-50. "Iterative" represents the iterative feedback regressor.}
\label{tab:ablation}
\end{table}

\subsubsection{The effectiveness of STE and KTD}

The upper part of Table \ref{tab:ablation} verifies the effectiveness of our proposed STE and KTD. Compared with CNN encoder+Iterative decoder, STE and KTD brings 4.7 and 1.3 mm improvement in PA-MPJPE metric respectively. Moreover, STE and KTD together further improves the performance by 6.5 mm. This proves the attention at different levels extracted by STE and KTD effectively complement rather than conflict each other.
% With Iterative and KTD decoder, the STE achieve 3.8 and 5.2 mm improvements in PA-MPJPE metric respectively. This demonstrates the temporal relation can lead to a better estimation in a continuous period of time, because we can use the information of the frames before and after to roughly infer the body pose of current frame. When using the CNN+STE encoder, the KTD outperforms Iterative decoder by 2.7mm in PA-MPJPE metric due to KTD's capability to pay more attention to the parent joint.

We can also observe that when using CNN encoder, the gain of KTD in PA-MPJPE metric is smaller than that when using CNN+STE encoder. Even there is a small decline in MPJPE metric. This is because the CNN loses too much spatial information due to the global pooling operation, and fails to provide detailed human body clue for KTD. However, with hard downsampling removed, STE not only preserves more spatial information, but also pay more attention to more informative locations, which makes KTD capture more precise attention between joints.
% proves our intuition that lots of spatial information is lost in global pooling operation. However, STE allocates reasonable attention on the feature map, which enables it to retain more spatial location information. The rich spatial information makes our KTD capture the relation between joints precisely.

% The encoder with only MSA-S can't capture temporal relation among frames, thus we equip it with MSA-T and MSA-C mechanism to make it a Spatial-Temporal Encoder (STE).
\subsubsection{Influence of different encoders} \label{ablation_ste}
In the middle part of Table \ref{tab:ablation}, we compare the performance of various forms of STE. SE denotes the encoder with only MSA-S. TE denotes the encoder with only MSA-T and CNN global pooling layer kept. STE$_{\text{parallel}}$v1 and STE$_{\text{parallel}}$v2 denote the Parallel Block w/o and w/ attentive addition respectively. We conclude that all the variants of STE benefit the model, while STE$_{\text{parallel}}$v2 yields the most significant gain. This is because the attentive weights dynamically computed
% computed for the two branches
in the Parallel Block effectively act as a valve which adjusts the proportion of temporal and spatial information passing through the network.
When it comes to occlusion or ambiguity, the valve will allow more temporal information to pass through to complement the lack of information in current frame, and do otherwise when the current frame is clear. Surprisingly, STE$_{\text{coupling}}$ yields only modest improvement over encoder with only MSA-S (49.8$\rightarrow$49.3), which has no temporal modeling capability. We also observe that STE$_{\text{coupling}}$ converges more slowly compared to other STE variants. We argue that flattening the spatial and temporal dimension together may harm human pose estimation mainly due to the extremely long sequence. 
%length when computing self-attention.
Tremendous irrelevant patches (such as background and joints that are too far apart) 
% flood input sequence and
overwhelm valid information, making it challenging for the current patch to allocate reasonable attention.
% Irrelevant patches (such as background and joints that are too far apart) flooding the patch sequence overwhelms valid information, making it challenging for the current patch to allocate reasonable attention. 

\subsubsection{Influence of different decoders} 
% Skeleton Topology Decoder (KTD) aims at explicitly modeling the topology relation between different joints. Specifically, it estimates the pose parameters from top to down along the kinematic tree.
We choose CNN+STE as the encoder and report the results with different decoders in the lower part of Table \ref{tab:ablation}. KTD$_{\text{random}}$ denotes the KTD on a randomly generated kinematic tree. KTD$_{\text{reverse}}$ denotes the KTD on the reverse kinematic tree, namely, exchange the relationship between parent joint and its children. Decoder$_{\text{vanilla}}$ denotes the standard decoder in \cite{transformer} with 6 layers. It takes as input the zero sequence of length 37 (24 for pose, 10 for shape and 3 for camera) and outputs SMPL parameters. %Moreover, it adopts the learnable positional encoding. 
We observe that KTD outperforms Iterative by a large margin. While KTD$_{\text{random}}$ and KTD$_{\text{reverse}}$ have no obvious improvement, even are slightly worse, proving unreasonable kinematic tree is useless prior knowledge, which brings difficulties to the optimization of the network. %It also demonstrates that correct kinematic topology relation can indeed help to the pose estimation. 
We also observe that Decoder$_{\text{vanilla}}$ brings no improvement. Although it can capture the relation between different joints with the self-attention mechanism, the predictions of all joints are generated simultaneously, not in the sequential way as KTD. As a result, it can not pay more attention to the parent joints.

\subsection{Visualization Analysis}

Figure \ref{fig:visual} includes qualitative results of MAED from two representative scenarios. 
%Neither the extreme pose in Figure \ref{fig:visual}(a) nor the cluttered background and occlusion in Figure \ref{fig:visual}(b) degrade the performance of our model, which predicts reasonable spatial and temporal attention maps. 
For these challenging cases including extreme pose in Figure \ref{fig:visual}(a) and cluttered background and occlusion in Figure \ref{fig:visual}(b), our model predicts reasonable spatial and temporal attention maps and further produce proper estimations.
%This proves the adaptability of our model to complex scenes. In addition, despite the back views in Figure \ref{fig:visual}(b) that further increases difficulties, our method yields smooth and accurate results.
% predict stable accurate results.

% which shows that our model does not overfit the front of human body.

%------------------------------------------------------------------------
\section{Conclusion}
This paper describes MAED, an approach that utilizes multi-level attentions at spatial-temporal level and human joint level for 3D human shape and pose estimation. We design multiple variants of MSA and STE Block to construct STE to learn spatial-temporal attention from the output feature of CNN backbone. In addition, we propose KTD, which simulates the process of joint rotation based on SMPL kinematic tree to decode human pose. 
MAED makes significant accuracy improvement on multiple datasets but also brings non-negligible computation overhead, which we explore further in the Sup. Mat.
Thus, future work could consider reducing computation overhead or extending this method to capture the relation between multiple people. %or recover human shape and pose on extreme occlusion scenario.

% \clearpage

{\small
\bibliographystyle{ieee_fullname}
\bibliography{egbib}
}

\end{document}